\documentclass[letterpaper, 10 pt, conference]{ieeeconf}  

\IEEEoverridecommandlockouts                              

\usepackage{graphicx}
\usepackage{amsmath} 
\usepackage{booktabs}
\usepackage{tabularx}
\usepackage{xcolor}
\usepackage{multirow}
\makeatletter
\let\NAT@parse\undefined
\makeatother

\usepackage[
    colorlinks=true,
    linkcolor=blue,
    citecolor=blue,
    urlcolor=blue,
    filecolor=blue,
    linktoc=all
]{hyperref}
 
\title{\LARGE \bf
Steeringless Drifting: Differential-Torque Control of a Four-Wheel Independently Driven Vehicle}

\author{Sheng Zhao$^{1,2}$, Zexin Wu$^{1}$, Dongyang Zhou$^{1}$, Bolin Zhao$^{2}$ and Xiaodong Wu$^{1}$
\thanks{*This work was supported by the National Science Foundation Grant 52472413. \textit{(Corresponding author: Xiaodong Wu)}}
\thanks{$^{1,2}$Sheng Zhao is with the School of Mechanical Engineering, Shanghai Jiao Tong University, Shanghai 200240, China, and also with the School of Mechanical and Aerospace Engineering, Nanyang Technological University, Singapore 639798, Singapore.
        {\tt\small zhaoshengkim@sjtu.edu.cn}}%
\thanks{$^{1}$Zexin Wu, Dongyang Zhou and Xiaodong Wu are with the School of Mechanical Engineering, Shanghai Jiao Tong University, Shanghai 200240, China.}%
\thanks{$^{2}$Bolin Zhao is with the School of Mechanical and Aerospace Engineering, Nanyang Technological University, Singapore 639798, Singapore.}%
}

\begin{document}

\maketitle
\thispagestyle{empty}
\pagestyle{empty}

\begin{abstract}

Control methods for emerging vehicle chassis architectures are important for autonomous driving near handling limits. Unlike conventional drift control, which relies on mechanical steering and rear-tire saturation, a steering-free four-wheel independently driven (4WID) vehicle can generate direct yaw moment through differential wheel torques. This paper proposes a differential-torque drift control method for such a vehicle. A double-track vehicle model incorporating four-wheel differential actuation is established, based on which a drift-equilibrium calculation method and a closed-loop drift controller are developed. The proposed approach is validated through simulations and experiments on a 1:10-scale vehicle. The results show that the vehicle can achieve steady circular drifting with a sideslip angle of approximately 20$^\circ$ and perform figure-eight drift tracking. This study demonstrates the feasibility of drift control using only differential wheel torques and provides a new perspective on near-limit control for steering-free vehicle architectures.
\end{abstract}

\section{INTRODUCTION}
Vehicle drifting is a deliberately sustained limit-handling condition characterized by a large body sideslip angle and operation of the tires near or beyond their adhesion limits. Unlike conventional stability control, which suppresses sideslip, autonomous drift control must stabilize an open-loop unstable operating condition while managing strongly coupled and saturated tire forces. It therefore provides both a demanding benchmark for nonlinear vehicle control and a means of extending the maneuverability envelope of autonomous vehicles~\cite{Wu_2026_vehicle}. Beyond motorsport and driving enjoyment, controlled drifting can create motion capabilities that are unavailable under conventional handling constraints, including rapid heading change for emergency collision avoidance~\cite{Zhao_2021_collision,Li_2023_collision,Zhao_2024_obstacle} and compact drift parking~\cite{Jelavic_2017_parking}. The analysis and control of drift equilibria are consequently relevant not only to the understanding of vehicle dynamics at the handling limits, but also to the development of future active-safety functions.

Early model-based studies formulated steady-state drifting as the stabilization of nonlinear equilibria of rear-wheel-drive vehicles and coordinated front-wheel steering with rear-wheel drive torque~\cite{Velenis_2011_steady,Hindiyeh_2014_controller}. This paradigm was subsequently extended from equilibrium regulation to general-path and transient drift maneuvers through feedback and mixed open-/closed-loop control~\cite{Goh_2020_beyond,Zhang_2018_drift}. Model predictive control (MPC) further enabled nonlinear dynamics, actuator limits, multiple equilibria, and variations in road friction to be incorporated explicitly into the controller~\cite{Bellegarda_2022_dynamic,Czibere_2021_mpc}. In parallel, reinforcement learning and data-driven methods reduced dependence on an exact tire model and demonstrated high-speed and consecutive drift maneuvers on scaled vehicles~\cite{Cutler_2016_autonomous,Cai_2020_highspeed,Lu_2023_consecutive}. More recently, model-based structure and learning have been combined to improve adaptability near the handling limits~\cite{Zhou_2025_learning}. Despite these advances, the prevailing actuation architecture remains centered on mechanical steering for lateral-force regulation and rear-wheel drive or braking for longitudinal-speed control and rear-tire saturation.

\begin{figure}[t] 
\centering 
\includegraphics[width=0.9\columnwidth]{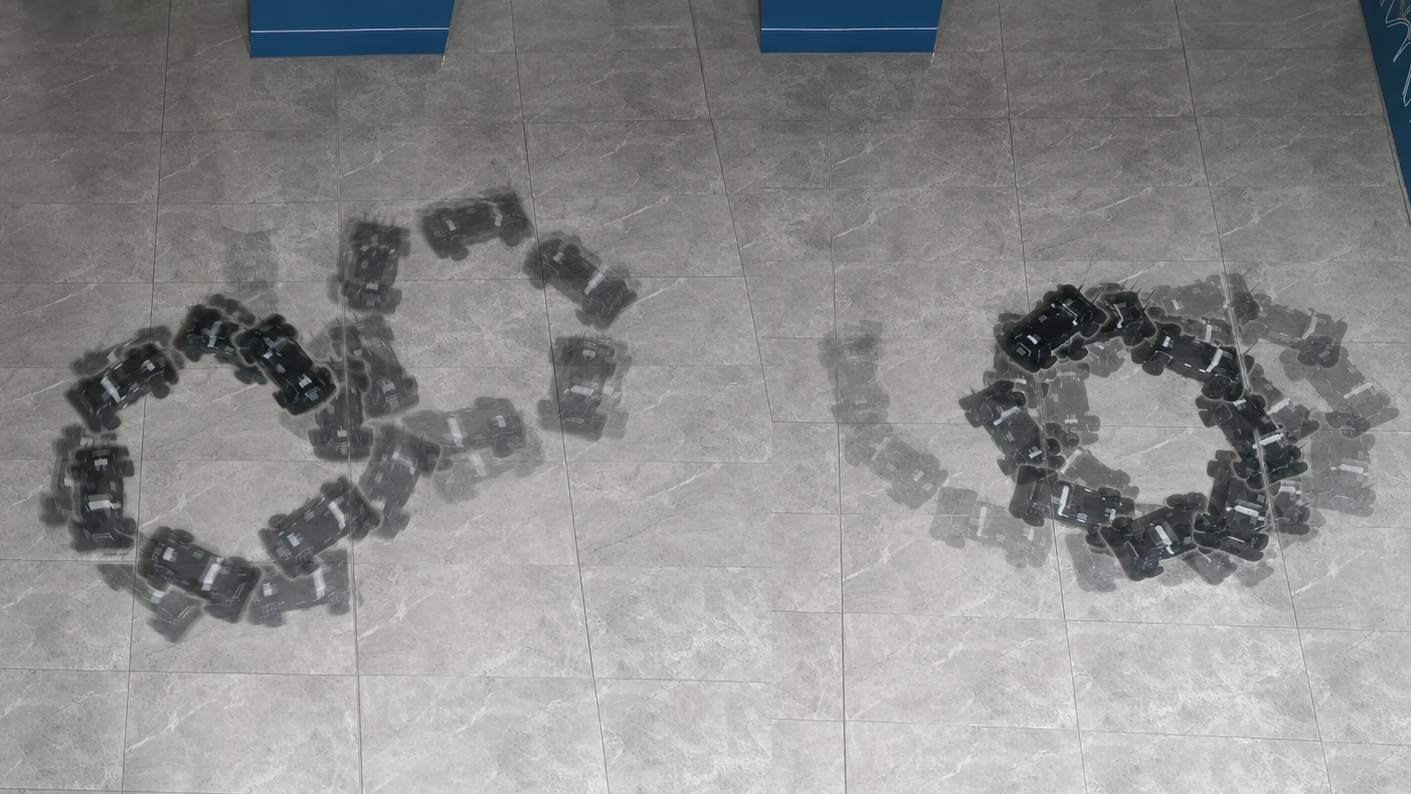} \caption{Time-lapse visualization of the RC vehicle executing a drift maneuver, with overlaid vehicle poses illustrating its motion trajectory. The corresponding experimental video is available at \href{https://youtu.be/ufDGAEhkPcU} {\textcolor{blue}{https://youtu.be/ufDGAEhkPcU}}.} 
\label{fig:RCdrift} 
\end{figure}

The development of intelligent electric chassis has introduced additional actuation dimensions. Four-wheel independent drive permits direct yaw-moment generation and flexible tire-force allocation, whereas active rear steering and four-wheel steering increase lateral control authority. These capabilities have been exploited in differential-steering control for 4WID vehicles~\cite{Tian_2018_differential}, automated drifting of 4WD-4WS and 4WD platforms~\cite{Xiao_2025_autonomousa,Zhao_2025_Automated}, drift control using torque vectoring~\cite{Lenzo_2024_torque}, and trajectory-following control for 4WID vehicles~\cite{Li_2022_trajectory}. Nevertheless, the drift controllers in these studies either retain a mechanical steering input or coordinate wheel torques with a steering subsystem. The steering and propulsion systems therefore remain jointly involved in generating and stabilizing the drift motion.

A steering-free 4WID chassis presents a fundamentally different control problem. With the wheel heading angles mechanically fixed, the vehicle generates its directly commanded yaw moment solely from the longitudinal-force difference between the left- and right-hand wheels. This architecture removes the conventional steering linkage, simplifies the mechanical layout, and reserves additional packaging space for drive, braking, and suspension components. It also invalidates a direct transfer of conventional drift-control methods. For a steered vehicle, a drift equilibrium is commonly obtained by solving the steady-state nonlinear vehicle equations for a prescribed speed and cornering condition, with the front-wheel angle orienting the front lateral force and the rear drive or brake torque regulating combined-slip saturation. The steering angle consequently shifts the attainable equilibrium sideslip angle and provides a direct stabilizing input. In a steering-free 4WID vehicle, by contrast, differential torque simultaneously determines the direct yaw moment, the total longitudinal force, and the combined-slip operating point of each tire. Drift initiation and stabilization must therefore be achieved through four coupled wheel torques, which requires a dedicated equilibrium formulation and control strategy.

To the best of the authors' knowledge, this paper presents the first complete drift-control framework for a steering-free 4WID vehicle. Figure~\ref{fig:RCdrift} illustrates the vehicle performing a figure-eight drift maneuver. The main contributions are summarized as follows:

\begin{itemize}

\item A double-track dynamic model and a drift-equilibrium calculation method are developed for the steering-free 4WID configuration. The formulation explicitly accounts for differential-torque yaw-moment generation and the coupling between the four wheel torques and nonlinear tire forces, thereby establishing the basis for drift control without a mechanical steering input.

\item A pulse--hold differential-torque controller is proposed to initiate and sustain drift using only the four wheel torques. Simulation and experiments on a 1:10-scale vehicle demonstrate steady circular drifting with a sideslip angle and figure-eight drift tracking, validating the feasibility of the proposed steering-free drift-control approach.

\end{itemize}

\section{Dynamic Modelling of the Steering-Free 4WID Vehicle}

\subsection{Vehicle Dynamics and Differential-Drive Mechanism}

\begin{figure}[t]
    \centering
    \includegraphics[width=0.8\columnwidth]{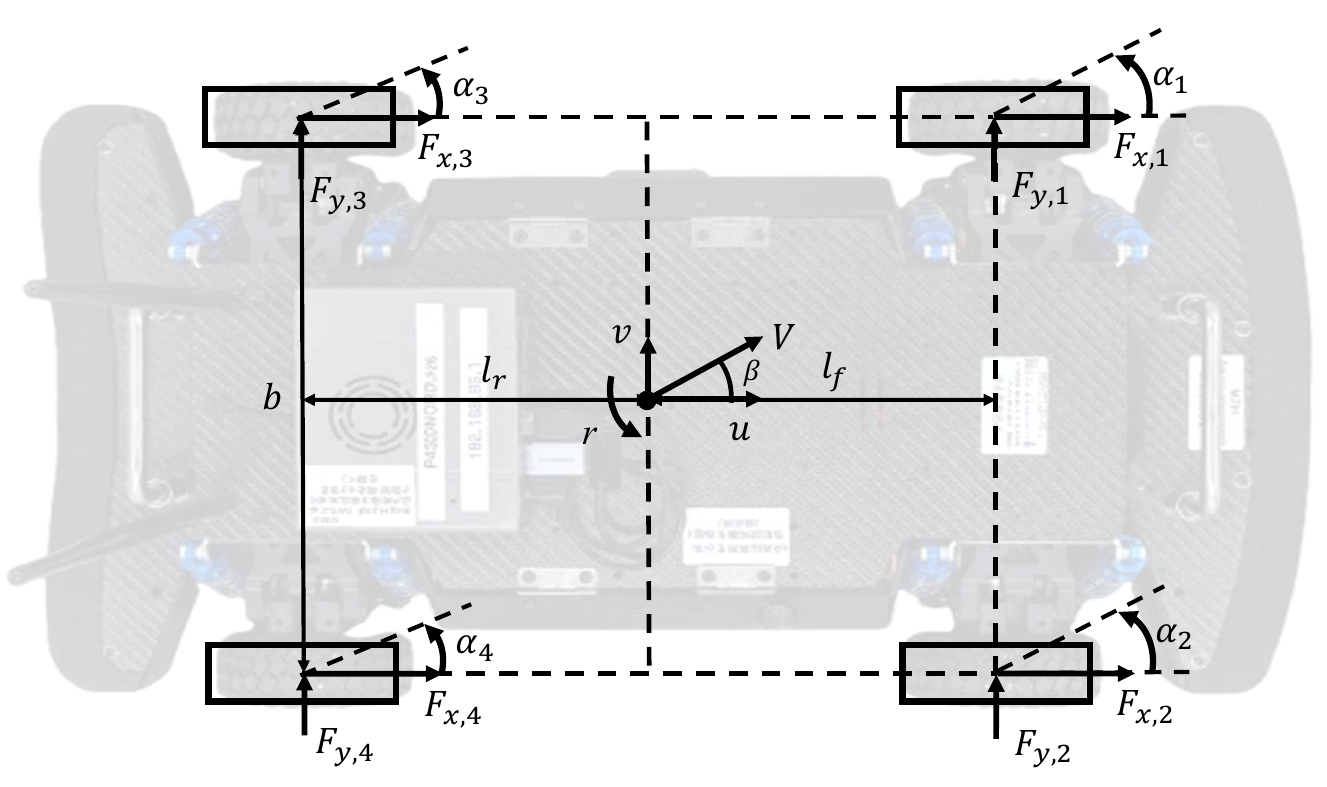}
    \caption{Planar double-track model and sign conventions of the
    steering-free 4WID vehicle. The force arrows indicate the positive
    directions of the corresponding body-frame components.}
    \label{fig:vehicle_model}
\end{figure}

Consider the planar steering-free 4WID vehicle illustrated in
Fig.~\ref{fig:vehicle_model}, whose wheels are numbered as front-left (FL),
front-right (FR), rear-left (RL), and rear-right (RR), corresponding to
wheels 1--4, respectively. A body-fixed frame $\{B\}$ is attached to the
vehicle center of mass (CoM), with its $x$-axis pointing forward and its
$y$-axis pointing to the left. Let
\begin{equation}
 \boldsymbol{x}=[X,\;Y,\;\psi,\;u,\;v,\;r]^{\mathrm T}
\end{equation}
denote the vehicle state, where $(X,Y)$ and $\psi$ are the global CoM
position and heading angle, respectively, and $u$, $v$, and $r$ are the
longitudinal velocity, lateral velocity, and yaw rate expressed in
$\{B\}$. Positive yaw is counterclockwise. The four wheel centers are
located at
\begin{equation}
 (x_i,y_i)\in
 \{(l_f,b/2),(l_f,-b/2),(-l_r,b/2),(-l_r,-b/2)\},
 \label{eq:wheel_positions}
\end{equation}
where $l_f$ and $l_r$ are the distances from the CoM to the front and
rear axles, and $b$ is the track width.

Suspension motion, load transfer, and pitch and roll dynamics are
neglected. The resulting three-degree-of-freedom rigid-body model is
\begin{flalign}
& \hspace{3em}\dot X = u\cos\psi-v\sin\psi, && \nonumber\\
& \hspace{3em}\dot Y = u\sin\psi+v\cos\psi, && \nonumber\\
& \hspace{3em}\dot\psi = r, && \label{eq:kinematics}\\
& \hspace{3em}m(\dot u-vr)
  = \sum_{i=1}^{4}F_{x,i},
  && \label{eq:long_dynamics}\\
& \hspace{3em}m(\dot v+ur)
  = \sum_{i=1}^{4}F_{y,i},
  && \label{eq:lat_dynamics}\\
& \hspace{3em}I_z\dot r
  = \sum_{i=1}^{4}
    \left(x_iF_{y,i}-y_iF_{x,i}\right),
  && \label{eq:yaw_dynamics}
\end{flalign}
Here, $m$ and $I_z$ are the vehicle mass and yaw moment of inertia,
respectively, and $F_{x,i}$ and $F_{y,i}$ are the longitudinal and
lateral tire forces in the body frame. 

Because all wheel steering angles are mechanically fixed at zero, yaw
motion is generated by four independently commanded motor torques. In
the unsaturated region, $F_{x,i}=\tau_i/R_w$, where $R_w$ is the wheel
radius. The longitudinal-force contribution to the yaw moment can
therefore be written as
\begin{equation}
 M_{z,x}=\frac{b}{2R_w}
 \left[(\tau_{\mathrm{FR}}+\tau_{\mathrm{RR}})
       -(\tau_{\mathrm{FL}}+\tau_{\mathrm{RL}})\right].
 \label{eq:differential_yaw}
\end{equation}
Equation~\eqref{eq:differential_yaw} reveals the essential control
authority of the steering-free chassis: the common-mode component of
the wheel torques primarily regulates propulsion, whereas their
left--right differential component produces the yaw moment normally
generated by a steering system. During drifting, the lateral forces in
\eqref{eq:yaw_dynamics} also contribute substantially to the total yaw
moment; hence, a simple kinematic differential-drive model is
insufficient for controller design.

The main variables and parameters used in the steering-free 4WID
vehicle model are summarized in Table~\ref{tab:vehicle_model_parameters}.

\begin{table}[t]
\centering
\caption{Vehicle-model symbols and parameters}
\label{tab:vehicle_model_parameters}
\renewcommand{\arraystretch}{1.10}
\begin{tabularx}{\columnwidth}{
    >{\centering\arraybackslash}p{0.16\columnwidth}
    X
    >{\centering\arraybackslash}p{0.15\columnwidth}}
\toprule
Symbol & Description & Unit \\
\midrule
$X,Y$ & Global CoM position & m \\
$\psi$ & Vehicle heading angle & rad \\
$u,v$ & Body-frame longitudinal and lateral velocities & m/s \\
$r$ & Yaw rate & rad/s \\
$m$ & Vehicle mass & kg \\
$I_z$ & Yaw moment of inertia & kg\,m$^2$ \\
$l_f,l_r$ & CoM-to-axle distances & m \\
$b$ & Track width & m \\
$R_w$ & Wheel radius & m \\
$x_i,y_i$ & Position of the $i$th wheel center & m \\
$F_{x,i}$ & Longitudinal tire force & N \\
$F_{y,i}$ & Lateral tire force & N \\
$\tau_i$ & Wheel driving torque & N\,m \\
$M_{z,x}$ & Differential-torque-induced yaw moment & N\,m \\
\bottomrule
\end{tabularx}
\end{table}

\subsection{Tire-Force and Actuator Modelling}

The velocity at the $i$th tire contact point is
\begin{equation}
 u_i=u-r y_i,\qquad v_i=v+r x_i .
 \label{eq:wheel_velocity}
\end{equation}
With zero steering angle, its slip angle is evaluated as
\begin{equation}
 \alpha_i=\operatorname{atan2}
 \left(v_i,\max(|u_i|,v_\epsilon)\right),
 \label{eq:slip_angle}
\end{equation}
where $v_\epsilon$ prevents numerical singularity at low speed. A
friction-circle-constrained tire model is employed to retain the
longitudinal--lateral coupling that is critical in a drift maneuver.
With the static normal load $F_{z,i}=mg/4$, the longitudinal force is
\begin{equation}
 F_{x,i}=\operatorname{sat}_{[-\eta\mu F_{z,i},\,\eta\mu F_{z,i}]}
 \left(\frac{\tau_i}{R_w}\right),
 \label{eq:long_tire_force}
\end{equation}
where $\mu$ is the tire--road friction coefficient and $\eta=0.995$
maintains a small numerical margin from the friction boundary. The
available lateral-force magnitude and the resulting lateral force are
\begin{align}
 \bar F_{y,i}&=\sqrt{\max\!\left[
  (\mu F_{z,i})^2-F_{x,i}^2,\,0\right]}, \label{eq:fy_capacity}\\
 F_{y,i}&=\operatorname{sat}_{[-\bar F_{y,i},\,\bar F_{y,i}]}
          (-C_{\alpha,i}\alpha_i), \label{eq:lat_tire_force}
\end{align}
where $C_{\alpha,i}$ is the cornering stiffness. The corresponding tire
utilization,
\begin{equation}
 \rho_i=\frac{\sqrt{F_{x,i}^2+F_{y,i}^2}}{\mu F_{z,i}},
 \label{eq:tire_utilization}
\end{equation}
is used below to penalize operating points that are unnecessarily close
to the friction limit.

The motor torque does not instantaneously follow its command. Its
dynamics are represented by a first-order model with both magnitude and
rate constraints,
\begin{align}
 \dot\tau_i&=\operatorname{sat}_{[-\dot\tau_{\max},\,\dot\tau_{\max}]}
 \left(\frac{\tau_{c,i}-\tau_i}{T_m}\right), \label{eq:motor_dynamics}\\
 |\tau_i|&\leq \tau_{\max}, \label{eq:motor_limit}
\end{align}
where $\tau_{c,i}$ is the commanded torque and $T_m$ is the actuator time
constant.The
vehicle sideslip angle reported in the simulation is
\begin{equation}
 \beta=\operatorname{atan2}(v,u).
 \label{eq:sideslip}
\end{equation}

\section{Drift Equilibrium and Control Algorithm Design}
\begin{figure*}[t] 
\centering
\includegraphics[width=\textwidth]{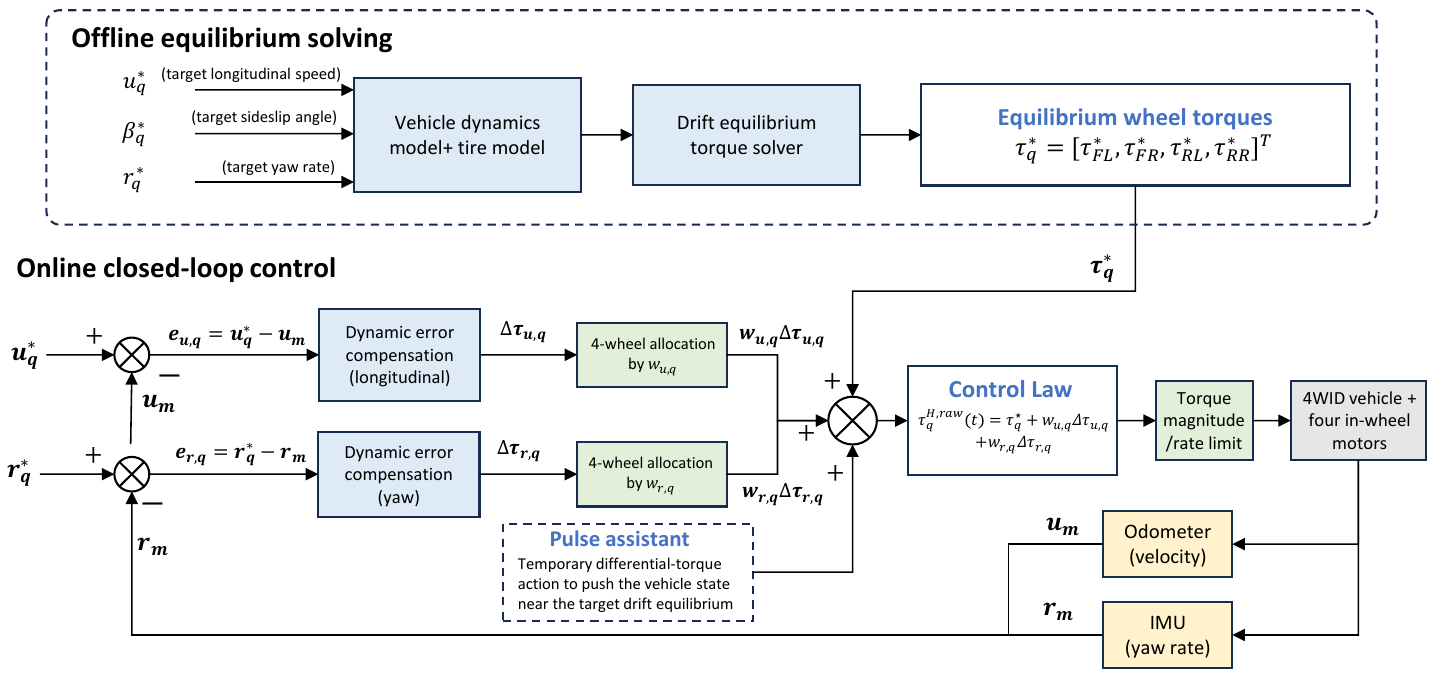} \caption{Overall architecture of the proposed steering-free 4WID drift control method. The desired longitudinal velocity, sideslip angle, and yaw rate are used to determine the equilibrium wheel torques offline. During online operation, longitudinal-velocity and yaw-rate feedback generate four-wheel torque corrections that are superimposed on the equilibrium torques. The optional differential-torque pulse is used only for drift initiation or state transition, while the sideslip angle is excluded from the online feedback loop.} 
\label{fig:control_framework} 
\end{figure*}

The overall architecture of the proposed steering-free drift control method is illustrated in Fig.~\ref{fig:control_framework}. The control scheme consists of an offline equilibrium-solving stage and an online closed-loop regulation stage. In the offline stage, the desired longitudinal velocity $u_q^\star$, sideslip angle $\beta_q^\star$, and yaw rate $r_q^\star$ are specified to determine the equilibrium wheel-torque vector $\boldsymbol{\tau}_q^\star$ using the vehicle dynamics and tire models. During online operation, the longitudinal-velocity and yaw-rate tracking errors are processed through two dynamic compensation channels. The resulting scalar torque corrections are distributed among the four wheels and superimposed on the equilibrium torque vector. After torque-magnitude and torque-rate constraints are imposed, the resulting commands are applied to the four independent drive motors. It should be noted that the sideslip angle is not included in the online feedback loop; it is used only to define the desired drift equilibrium and may be monitored during vehicle operation.

\subsection{Differential-Torque Drift Equilibrium}

A steady drift is a relative equilibrium: the vehicle translates and
rotates in the inertial frame, while its body-frame longitudinal
velocity, lateral velocity, and yaw rate remain constant. Let
$q\in\mathcal{Q}$ index a desired steady operating condition, where
$\mathcal{Q}$ may contain nominal-driving and drift equilibria. Its
target body-frame state is parameterized by
\begin{equation}
 \boldsymbol{\xi}_q^\star
 = [u_q^\star,\;v_q^\star,\;r_q^\star]^{\mathrm T},
 \qquad
 v_q^\star=u_q^\star\tan\beta_q^\star ,
 \label{eq:equilibrium_v}
\end{equation}
where $u_q^\star$, $\beta_q^\star$, and $r_q^\star$ are design
parameters. At a steady operating condition, the motor dynamics in
\eqref{eq:motor_dynamics} imply
$\boldsymbol{\tau}_c^\star=\boldsymbol{\tau}^\star$. The equilibrium
torque vector
$\boldsymbol{\tau}=[\tau_{\mathrm{FL}},\tau_{\mathrm{FR}},
\tau_{\mathrm{RL}},\tau_{\mathrm{RR}}]^{\mathrm T}$ is therefore
obtained by enforcing
$\dot u=\dot v=\dot r=0$ in
\eqref{eq:long_dynamics}--\eqref{eq:yaw_dynamics}.

For compactness, define
$F_{x,i,q}(\boldsymbol{\tau})
=F_{x,i}(\boldsymbol{\xi}_q^\star,\boldsymbol{\tau})$ and
$F_{y,i,q}(\boldsymbol{\tau})
=F_{y,i}(\boldsymbol{\xi}_q^\star,\boldsymbol{\tau})$. The corresponding
resultant forces and yaw moment are
\begin{align}
\mathcal{F}_{x,q}(\boldsymbol{\tau})
&=\sum_iF_{x,i,q}(\boldsymbol{\tau}), \nonumber\\
\mathcal{F}_{y,q}(\boldsymbol{\tau})
&=\sum_iF_{y,i,q}(\boldsymbol{\tau}), \nonumber\\
\mathcal{M}_{z,q}(\boldsymbol{\tau})
&=\sum_i\!\left[x_iF_{y,i,q}(\boldsymbol{\tau})
-y_iF_{x,i,q}(\boldsymbol{\tau})\right].
\label{eq:equilibrium_wrench}
\end{align}
Consistent with the longitudinal model in \eqref{eq:long_dynamics},
which contains no additional rolling-resistance or aerodynamic-drag
term, the equilibrium residual is
\begin{equation}
\boldsymbol{g}_q(\boldsymbol{\tau})=
\begin{bmatrix}
 m^{-1}\mathcal{F}_{x,q}(\boldsymbol{\tau})
   +v_q^\star r_q^\star\\
 m^{-1}\mathcal{F}_{y,q}(\boldsymbol{\tau})
   -u_q^\star r_q^\star\\
 I_z^{-1}\mathcal{M}_{z,q}(\boldsymbol{\tau})
\end{bmatrix}.
\label{eq:equilibrium_residual}
\end{equation}
The first two components of \eqref{eq:equilibrium_residual} balance the
centripetal coupling terms in the body-fixed translational dynamics,
while the third enforces yaw-moment equilibrium.

The friction-circle saturation makes the equilibrium equations
nonlinear, and the four independent wheel torques make the allocation
nonunique. A particular equilibrium allocation
$\boldsymbol{\tau}_q^\star$ is selected from the feasible set by
\begin{equation}
\begin{split}
J_q(\boldsymbol{\tau})
={}&w_\tau
 \left\|\frac{\boldsymbol{\tau}}{\tau_{\max}}\right\|_2^2+w_\rho\sum_{i=1}^{4}
 \rho_{i,q}(\boldsymbol{\tau})^{\gamma_\rho}.
\end{split}
\label{eq:equilibrium_cost}
\end{equation}
The equilibrium allocation is consequently defined by
\begin{equation}
\begin{aligned}
\boldsymbol{\tau}_q^\star
&=\underset{\boldsymbol{\tau}}{\operatorname{arg\,min}}\quad
J_q(\boldsymbol{\tau})\\
\text{subject to}\quad
&\boldsymbol{g}_q(\boldsymbol{\tau})=\boldsymbol{0},\\
&|\tau_i|\leq\tau_{\max},\qquad i=1,\ldots,4 .
\end{aligned}
\label{eq:equilibrium_optimization}
\end{equation}
Here, $\rho_{i,q}$ is the tire utilization in
\eqref{eq:tire_utilization}; $w_\tau$ and $w_\rho$ are nonnegative
weights; and $\gamma_\rho>2$ determines how strongly near-saturation
solutions are penalized. The values used in the current implementation
are $w_\tau=0.25$, $w_\rho=0.50$, and $\gamma_\rho=6$. The first term
limits actuator effort, whereas the second preserves tire-force margin.

The constrained problem in \eqref{eq:equilibrium_optimization} is solved
using multistart sequential quadratic programming. Initial guesses
spanning common-mode and differential-mode torque distributions are
used to reduce sensitivity to local minima. The resulting
$\boldsymbol{\tau}_q^\star$ is stored as the feedforward torque for
holding operating condition $q$. For a laterally symmetric vehicle, an
opposite-turn solution may be initialized by exchanging the left- and
right-wheel entries and reversing the signs of
$(\beta_q^\star,r_q^\star)$; nevertheless, each candidate is checked
against \eqref{eq:equilibrium_residual} rather than assumed to remain
feasible.

\begin{figure}[ht]
\centering
\includegraphics[width=\columnwidth]{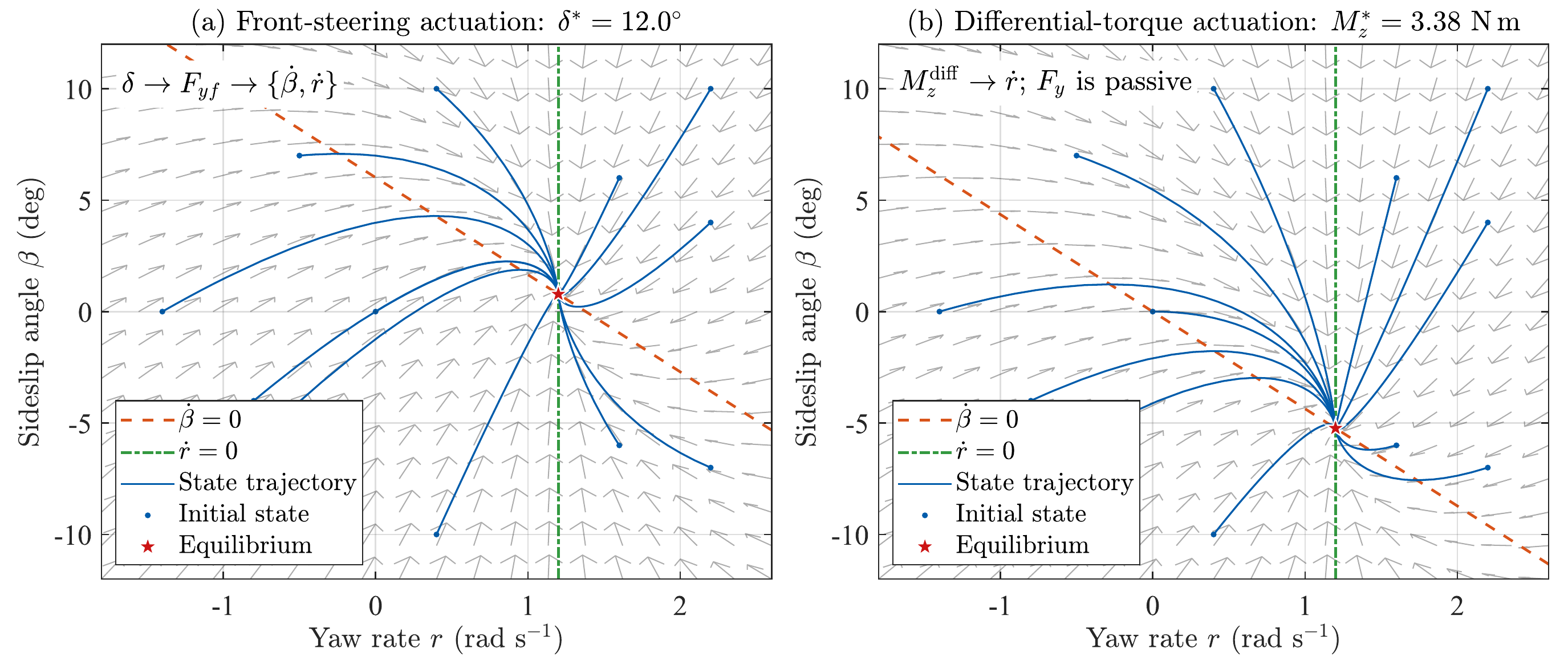}
\caption{Comparison of the $r$--$\beta$ phase planes for a
front-steered vehicle and a steering-free 4WID vehicle under the same
longitudinal speed and target yaw rate. The dashed and dash-dotted lines
denote the $\dot{\beta}=0$ and $\dot r=0$ nullclines, respectively, and
their intersection defines the equilibrium. The steering angle acts
through the front lateral tire force, whereas the differential wheel
torques generate a direct yaw moment.}
\label{fig:steering_4wid_phase_plane}
\end{figure}

The $r$--$\beta$ phase-plane comparison in
Fig.~\ref{fig:steering_4wid_phase_plane} further clarifies the actuation
mechanism underlying the equilibrium solution. In both architectures,
an equilibrium is located at the intersection of the
$\dot{\beta}=0$ and $\dot r=0$ nullclines. For the steering-free 4WID
vehicle, the existence of this intersection demonstrates that a steady
lateral--yaw operating condition can be sustained without a steering
angle: the differential longitudinal forces provide the required yaw
moment, while the lateral tire forces generated passively by tire slip
balance the lateral acceleration. The differential moment therefore
does not replace the lateral tire force; instead, it modifies the yaw
motion and thereby changes the slip angles and lateral-force
distribution. The input pathways are nevertheless fundamentally
different. A front steering angle acts on both $\dot{\beta}$ and
$\dot r$ through the front lateral force, whereas a differential yaw
moment enters $\dot r$ directly and influences $\beta$ only through the
coupled vehicle dynamics. Consequently, the two architectures can
produce different equilibrium sideslip angles and nullcline locations
at the same longitudinal speed and yaw rate. The convergent trajectories
in the linear phase planes indicate local attraction of the selected
equilibria for the adopted parameters; however, because a linear tire
model is used, the comparison establishes the local equilibrium
structure and actuation difference rather than proving the existence
or stability of a large-sideslip nonlinear drift.

\subsection{Pulse--Hold Drift Control}

Stabilization of a selected drift equilibrium is formulated as a
feedforward--feedback regulation problem. During the hold phase for
operating condition $q$, the equilibrium torque
$\boldsymbol{\tau}_q^\star$ obtained from
\eqref{eq:equilibrium_optimization} supplies the nominal control action,
while dynamic error compensation is constructed from the measured
longitudinal speed and yaw rate. Define
\begin{equation}
 e_{u,q}=u_q^\star-u_m,\qquad
 e_{r,q}=r_q^\star-r_m ,
\label{eq:tracking_errors}
\end{equation}
where $u_m$ is the measured body-frame longitudinal speed and $r_m$ is
the bias-corrected measured yaw rate. The corresponding integral states
are
\begin{equation}
 z_{u,q}(t)=\int_{t_q}^{t}e_{u,q}(\sigma)\,d\sigma,\qquad
 z_{r,q}(t)=\int_{t_q}^{t}e_{r,q}(\sigma)\,d\sigma ,
\label{eq:error_integral_states}
\end{equation}
where $t_q$ denotes the time at which the hold controller for condition
$q$ is activated.

The speed and yaw-rate correction torques are generated by
\begin{align}
\Delta\tau_{u,q}
&=\operatorname{sat}_{[-\bar\tau_u,\,\bar\tau_u]}
 \left(K_{P,u}e_{u,q}+K_{I,u}z_{u,q}
       -K_{D,u}\dot u_m\right),
\label{eq:speed_feedback}\\
\Delta\tau_{r,q}
&=\operatorname{sat}_{[-\bar\tau_r,\,\bar\tau_r]}
 \left(K_{P,r}e_{r,q}+K_{I,r}z_{r,q}
       -K_{D,r}\dot r_m\right),
\label{eq:yaw_feedback}
\end{align}
Here, the $K_P$, $K_I$, and $K_D$ coefficients weight the instantaneous
tracking error, accumulated error, and measured-rate feedback,
respectively, for the longitudinal and yaw channels. The measured-rate
terms are used in place of reference differentiation, thereby avoiding
an impulsive correction when the desired equilibrium is changed. The
integral states and correction torques are bounded to limit windup and
to prevent the feedback action from overriding the equilibrium
feedforward allocation.

Let $\boldsymbol{w}_{u,q}\in\mathbf{R}^4$ and
$\boldsymbol{w}_{r,q}\in\mathbf{R}^4$ denote the four-wheel allocation
vectors for the speed and yaw-rate corrections. The raw hold command is
\begin{equation}
\boldsymbol{\tau}_{q}^{\mathrm{H,raw}}(t)
=\boldsymbol{\tau}_q^\star
+\boldsymbol{w}_{u,q}\Delta\tau_{u,q}
+\boldsymbol{w}_{r,q}\Delta\tau_{r,q}.
\label{eq:hold_control}
\end{equation}
Thus, $\boldsymbol{\tau}_q^\star$ determines the nominal force and yaw
moment required by the drift equilibrium, while the two dynamic
compensation channels attenuate longitudinal-speed and yaw-rate
deviations. The
allocation vectors are controller parameters and can be selected using
the chassis geometry, actuator authority, and offline simulation. They
may vary with $q$ when different drift equilibria require different
control authority.

The desired sideslip angle $\beta_q^\star$ enters the offline
equilibrium calculation through \eqref{eq:equilibrium_v}, but it is not
used as an online feedback variable. Consequently, the implemented
controller is an open-$\beta$ equilibrium regulator: large sideslip is
encoded in $\boldsymbol{\tau}_q^\star$, whereas the closed-loop signals
are limited to the experimentally reliable measurements $u_m$ and
$r_m$. This avoids injecting errors from a sideslip estimate reconstructed
from IMU accelerations, yaw rate, and odometry speed.

Before being sent to the motor dynamics, the raw hold command is
magnitude- and rate-limited. At sampling instant $k$, define
$\delta_\tau=\dot\tau_{c,\max}\Delta t$, where $\Delta t$ is the control
period and $\dot\tau_{c,\max}$ is the allowable command slew rate. For
each wheel,
\begin{align}
\bar\tau_{c,i}[k]
&=\operatorname{sat}_{[-\tau_{\max},\,\tau_{\max}]}
  \left(\tau_{q,i}^{\mathrm{H,raw}}[k]\right),
\label{eq:command_magnitude_limit}\\
\tau_{c,i}[k]
&=\tau_{c,i}[k-1]
+\operatorname{sat}_{[-\delta_\tau,\,\delta_\tau]}
 \left(\bar\tau_{c,i}[k]-\tau_{c,i}[k-1]\right).
\label{eq:command_rate_limit}
\end{align}
The resulting $\boldsymbol{\tau}_c$ is applied through
\eqref{eq:motor_dynamics}. All feedback gains, correction bounds,
allocation vectors, and command limits are tunable parameters; their
numerical values are reported with the corresponding validation
scenario.

Because the equilibrium regulator is local, its feedforward torque may
not initiate the desired drift when the current vehicle state lies
outside the target equilibrium's attraction region. In engineering
practice, a short differential-torque pulse can therefore be applied
before the hold controller is enabled, followed by a smooth transfer to
\eqref{eq:hold_control}. The pulse only assists the state transition; it
does not replace or modify the feedback law that stabilizes the
resulting drift equilibrium.

\section{validation and results}

\subsection{Simulation setup implementation and validation results}

The proposed control framework was implemented and evaluated in MATLAB
R2023a. The nominal vehicle model was parameterized with
$m=7.0~\mathrm{kg}$,
$I_z=0.18~\mathrm{kg\,m^2}$,
$l_f=l_r=0.175~\mathrm{m}$,
$b=0.310~\mathrm{m}$,
$R_w=0.048~\mathrm{m}$,
$C_{\alpha,i}=46~\mathrm{N/rad}$,
$\mu=0.85$,
$v_\epsilon=0.08~\mathrm{m/s}$,
$T_m=0.035~\mathrm{s}$,
$\dot{\tau}_{\max}=5.0~\mathrm{N\,m/s}$, and
$\tau_{\max}=0.8~\mathrm{N\,m}$.
The nonlinear vehicle dynamics in
Eqs.~\eqref{eq:kinematics}--\eqref{eq:motor_limit}
were numerically integrated using a fixed-step fourth-order
Runge--Kutta method.

\begin{table*}[t]
\centering
\caption{Quantitative error metrics for the circular and figure-eight drift experiments.}
\label{tab:experimental_error_metrics}

\renewcommand{\arraystretch}{1.15}
\setlength{\tabcolsep}{3.5pt}

\resizebox{\textwidth}{!}{%
\begin{tabular}{lcccccccccccccc}
\toprule
\multirow{3}{*}{Experiment}
& \multicolumn{2}{c}{Yaw rate (rad/s)}
& \multicolumn{2}{c}{Sideslip angle ($^\circ$)}
& \multicolumn{2}{c}{Velocity (m/s)}
& \multicolumn{8}{c}{Motor torque ($\mathrm{N\,m}$)}
\\

\cmidrule(lr){2-3}
\cmidrule(lr){4-5}
\cmidrule(lr){6-7}
\cmidrule(lr){8-15}

& \multirow{2}{*}{Max.}
& \multirow{2}{*}{RMSE}
& \multirow{2}{*}{Max.}
& \multirow{2}{*}{RMSE}
& \multirow{2}{*}{Max.}
& \multirow{2}{*}{RMSE}
& \multicolumn{2}{c}{FL}
& \multicolumn{2}{c}{FR}
& \multicolumn{2}{c}{RL}
& \multicolumn{2}{c}{RR}
\\

\cmidrule(lr){8-9}
\cmidrule(lr){10-11}
\cmidrule(lr){12-13}
\cmidrule(lr){14-15}

& & & & & & 
& Max. & RMSE
& Max. & RMSE
& Max. & RMSE
& Max. & RMSE
\\

\midrule

Circular
& 0.0564 & 0.0254
& 0.8141 & 0.5407
& 0.0371 & 0.0156
& 0.0974 & 0.0255
& 0.1000 & 0.0301
& 0.0999 & 0.0212
& 0.0988 & 0.0175
\\

Figure-eight
& 0.1522 & 0.0579
& 5.2698 & 1.1929
& 0.3500 & 0.0627
& 0.0946 & 0.0277
& 0.0962 & 0.0281
& 0.0995 & 0.0194
& 0.0999 & 0.0204
\\

\bottomrule
\end{tabular}%
}
\end{table*}

\begin{figure}[t]
    \centering
    \includegraphics[width=\columnwidth]{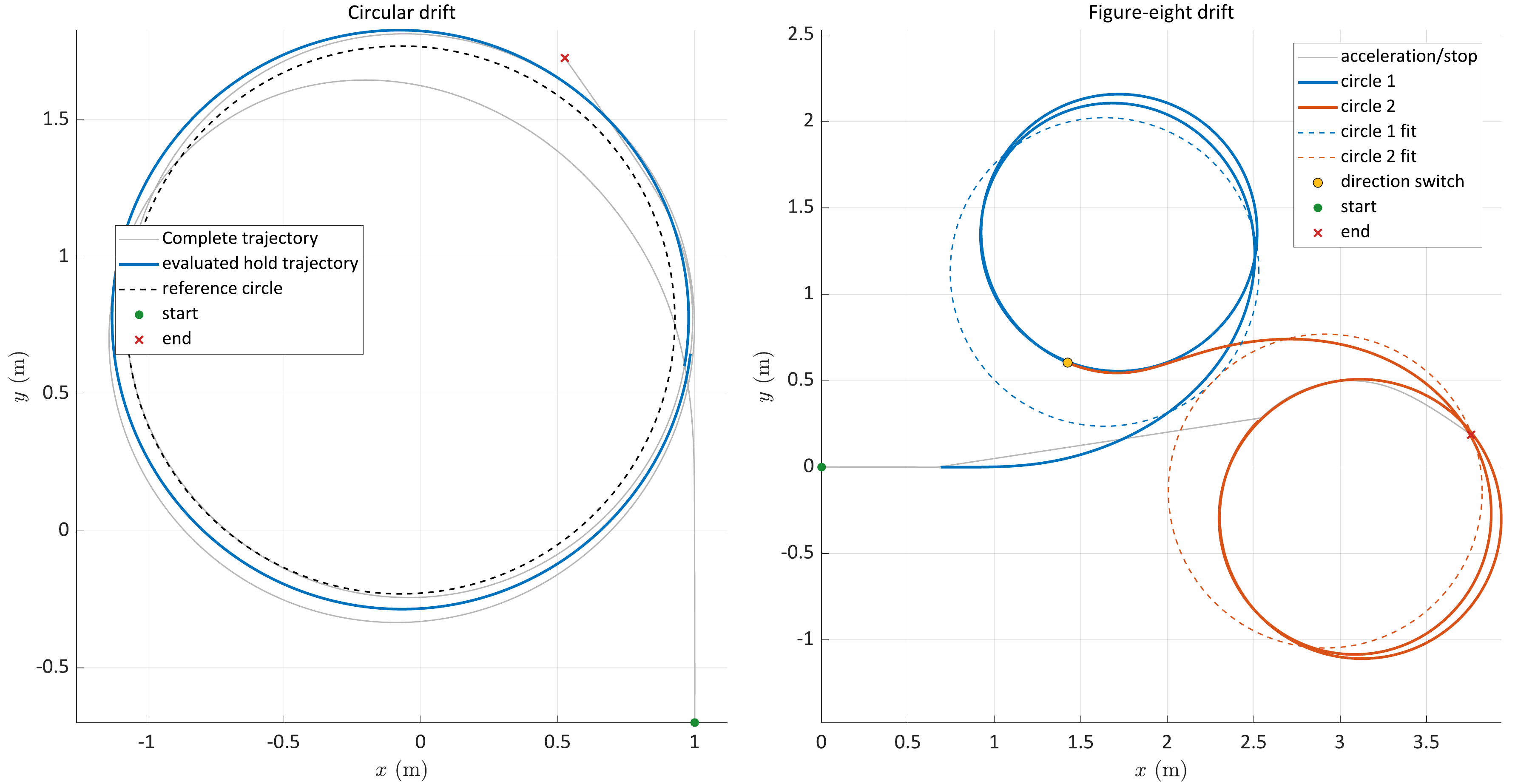}
    \caption{Vehicle trajectories in the two representative scenarios: (a) steady-state circular drifting and (b) figure-eight drifting.}
    \label{fig:traj}
\end{figure}

Two representative scenarios were considered: steady-state circular
drifting and figure-eight drifting. The circular drift scenario was used
to evaluate convergence to and stabilization around a single prescribed
drift equilibrium. In contrast, the figure-eight scenario involved
successive transitions between drift equilibria of opposite directions
and therefore provided a more demanding assessment of transient tracking
and continuous drift stabilization. The resulting vehicle trajectories
are shown in Fig.~\ref{fig:traj}, while the corresponding quantitative
tracking-error metrics are summarized in
Table~\ref{tab:experimental_error_metrics}.

As reported in Table~\ref{tab:experimental_error_metrics}, the circular
drift simulation achieved relatively small tracking errors, with RMSE
values of $0.0254~\mathrm{rad/s}$, $0.5407^\circ$, and
$0.0156~\mathrm{m/s}$ for yaw rate, sideslip angle, and
velocity, respectively. These results indicate that the controller can
accurately regulate the vehicle around a fixed drift equilibrium.
Larger transient errors were observed in the figure-eight scenario,
particularly in the sideslip angle and longitudinal velocity. This
increase is primarily associated with the rapid reversal of the desired
yaw motion and the transition between two drift equilibria of opposite
directions. Nevertheless, the corresponding RMSE values remained limited
to $0.0579~\mathrm{rad/s}$ for yaw rate, $1.1929^\circ$ for sideslip
angle, and $0.0627~\mathrm{m/s}$ for velocity. Moreover, the
maximum motor-torque tracking error remained at or below
$0.1~\mathrm{N\,m}$ for all four wheels in both scenarios, with RMSE
values below $0.031~\mathrm{N\,m}$. These results demonstrate that the
proposed controller provides satisfactory equilibrium tracking, actuator
coordination, and transition capability under both steady and
direction-reversing drift maneuvers.

\subsection{Real vehicle hardware platform}
\begin{figure}[ht]
    \centering
    \includegraphics[width=0.95\columnwidth]{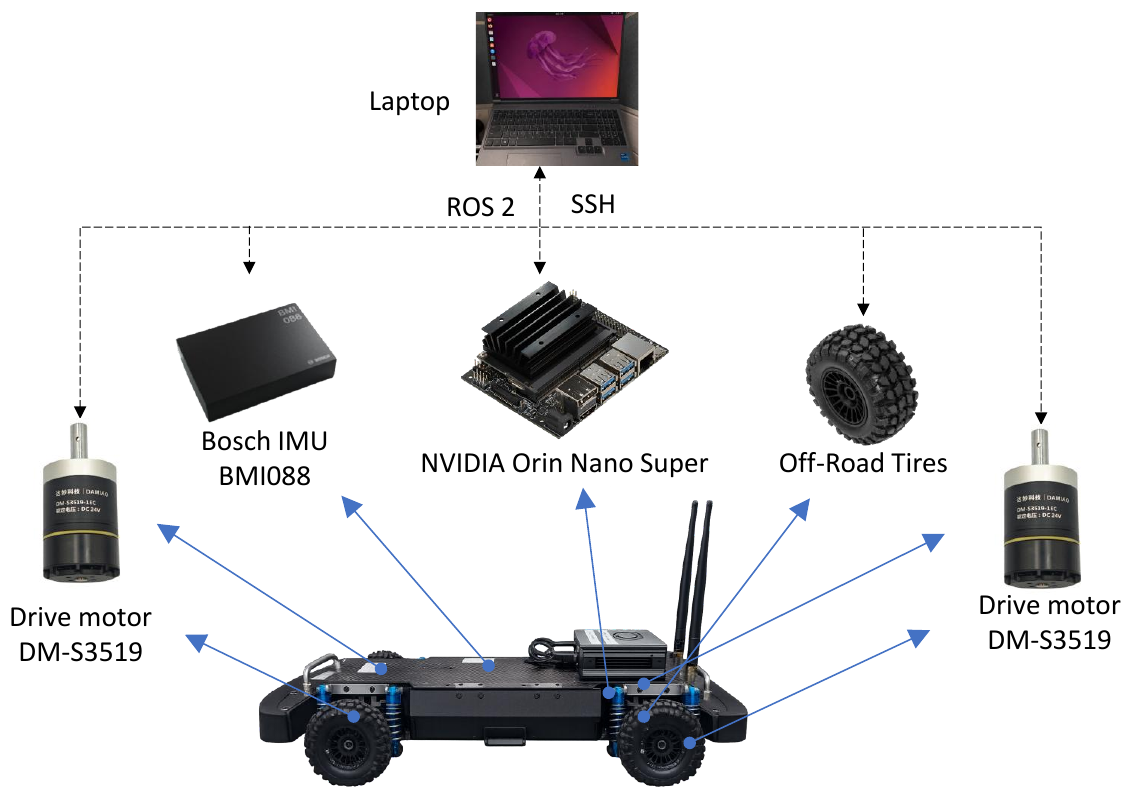}
    \caption{Hardware platform of the 1:10-scale four-wheel-independent-drive vehicle.}
    \label{fig:RChardware}
\end{figure}

A 1:10-scale 4WID vehicle was developed
to validate the proposed drift control method, as shown in
Fig.~\ref{fig:RChardware}. The vehicle is equipped with an NVIDIA Jetson
Orin Nano Super as the onboard computing unit, running Ubuntu 22.04 and
ROS~2. During the experiments, the onboard computer communicates with a
laptop through a wireless local-area network, and remote operation,
program deployment, and data monitoring are performed via Secure Shell
(SSH). The principal vehicle, actuator, and sensor parameters are
summarized in Table~\ref{tab:platform_parameters}.

\begin{table}[ht]
    \centering
    \caption{Key parameters of the 1:10-scale 4WID platform.}
    \label{tab:platform_parameters}
    \small
    \renewcommand{\arraystretch}{1.15}
    \begin{tabularx}{\columnwidth}{@{}Xcc@{}}
        \toprule
        Parameter & Symbol & Value \\
        \midrule

        Vehicle mass
        & $m$
        & $7.0~\mathrm{kg}$ \\

        Yaw moment of inertia
        & $I_z$
        & $0.18~\mathrm{kg\,m^2}$ \\

        Distance from CoM to front axle
        & $l_f$
        & $0.175~\mathrm{m}$ \\

        Distance from CoM to rear axle
        & $l_r$
        & $0.175~\mathrm{m}$ \\

        Track width
        & $b$
        & $0.310~\mathrm{m}$ \\

        Wheel radius
        & $R_w$
        & $0.048~\mathrm{m}$ \\

        Tire cornering stiffness
        & $C_{\alpha,i}$
        & $46~\mathrm{N/rad}$ \\

        Maximum motor torque rate
        & $\dot{\tau}_{\max}$
        & $5.0~\mathrm{N\,m/s}$ \\

        Maximum motor torque
        & $\tau_{\max}$
        & $1.2~\mathrm{N\,m}$ \\

        Accelerometer resolution
        & --
        & $0.09~\mathrm{mg}$ \\

        Gyroscope resolution
        & --
        & $0.004~\mathrm{deg/s}$ \\

        \bottomrule
    \end{tabularx}
\end{table}

\subsection{Real vehicle circular drift results}

\begin{figure}[t]
\centering
\includegraphics[width=0.9\columnwidth]{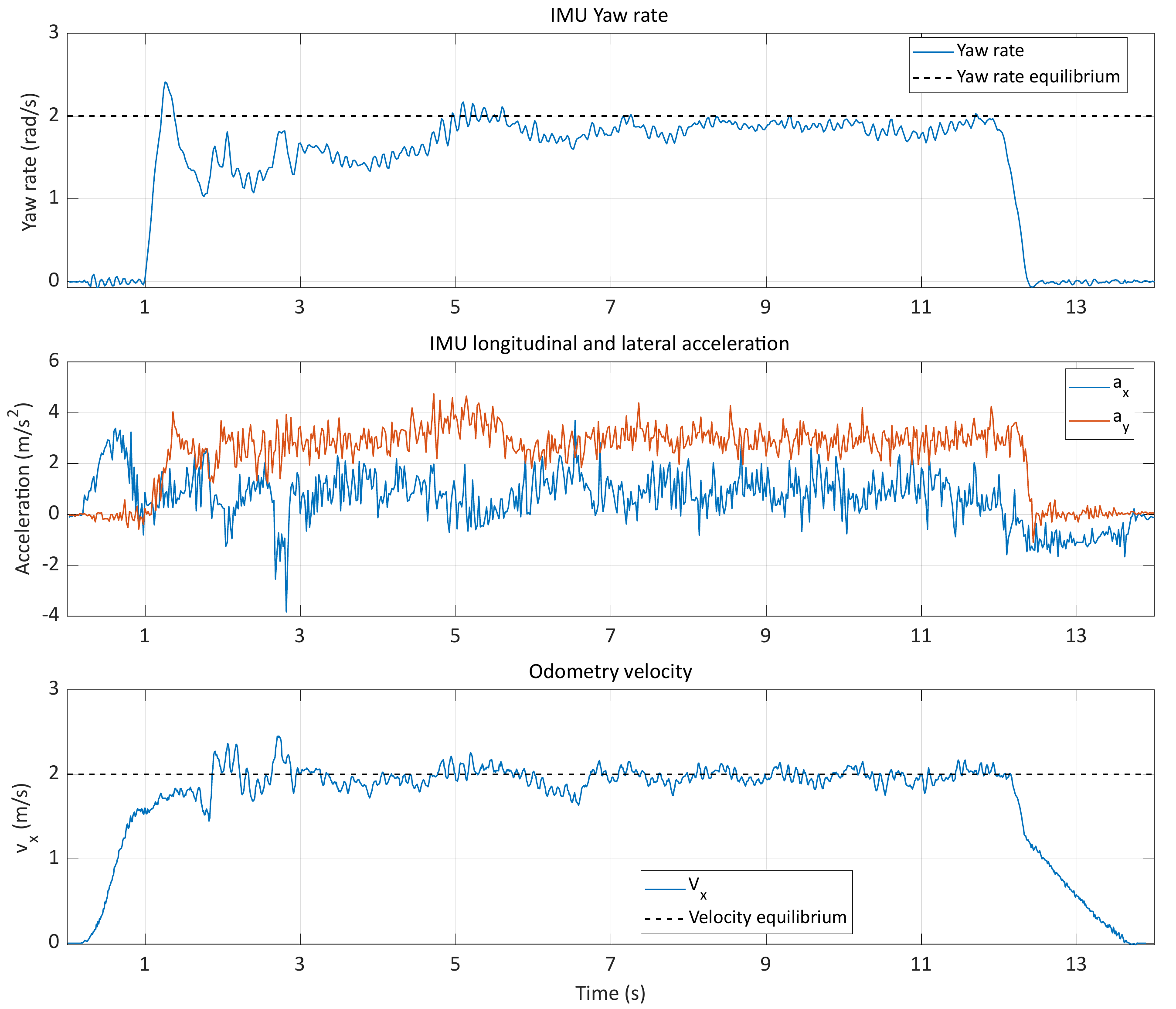}
\caption{Measured vehicle states during the circular drift experiment, including the yaw rate, longitudinal and lateral accelerations, longitudinal velocity.}
\label{fig:circle_via}
\end{figure}

\begin{figure}[t]
\centering
\includegraphics[width=0.9\columnwidth]{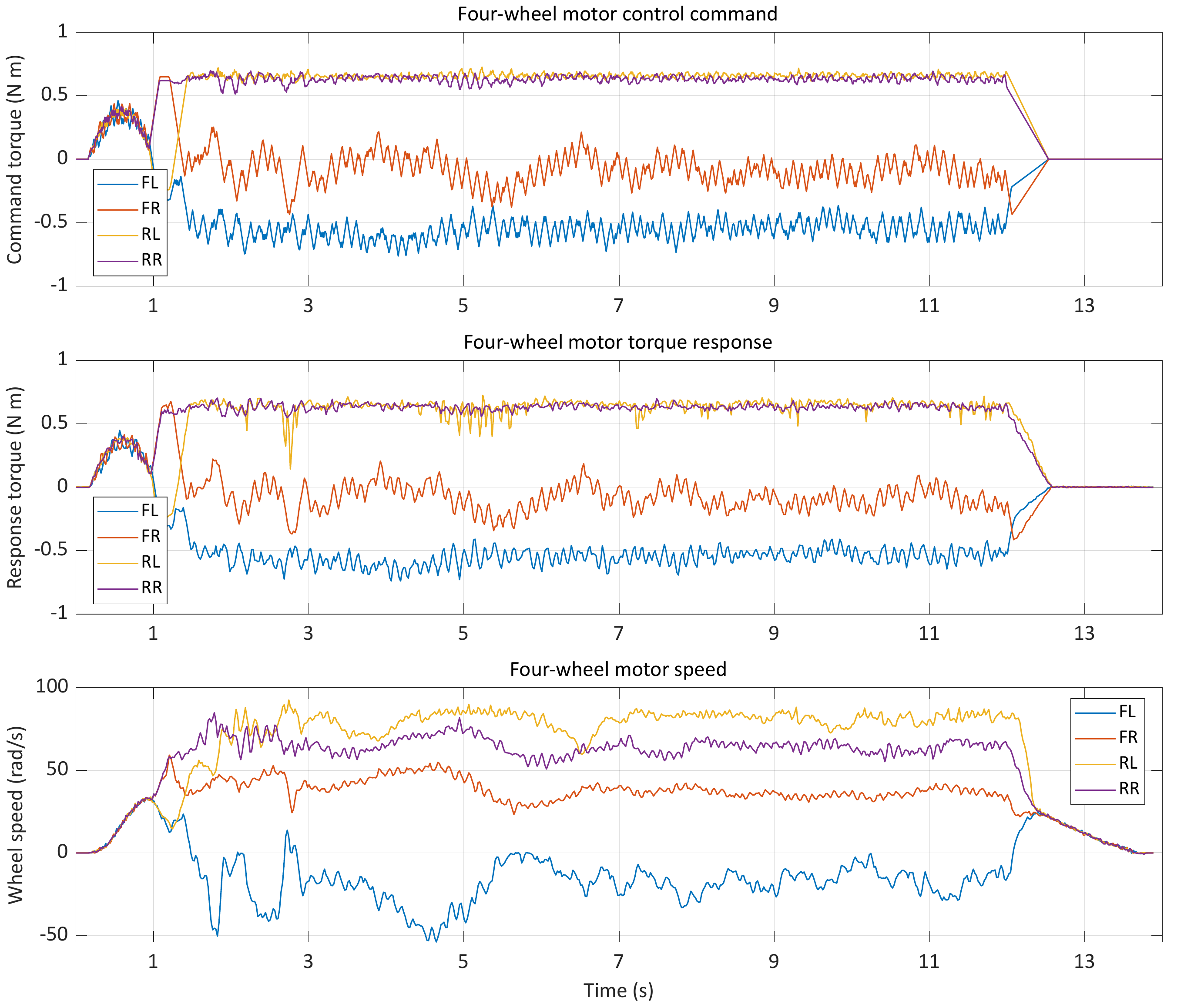}
\caption{Four-wheel torque commands, motor torque feedback, and wheel-speed responses during the circular drift experiment.}
\label{fig:circle_cmd}
\end{figure}
A steady-state circular drift experiment was first conducted to evaluate the capability of the proposed controller to initiate and maintain a prescribed drift equilibrium on the steering-free 4WID platform. The measured vehicle states are presented in Fig.~\ref{fig:circle_via}, while the corresponding four-wheel torque commands, motor torque feedback, and wheel-speed responses are shown in Fig.~\ref{fig:circle_cmd}.

During the experiment, the vehicle was first accelerated toward the prescribed longitudinal velocity. A differential-torque pulse was subsequently applied to generate the yaw moment required for drift initiation. After the vehicle approached the desired drift equilibrium, the wheel torques were continuously regulated to maintain the target yaw-rate response and longitudinal velocity. The results demonstrate that the proposed controller can successfully initiate the drift motion and sustain a stable circular drift without mechanical steering actuation.

\subsection{Real vehicle figure-eight drift results}
\begin{figure}[t]
\centering
\includegraphics[width=0.9\columnwidth]{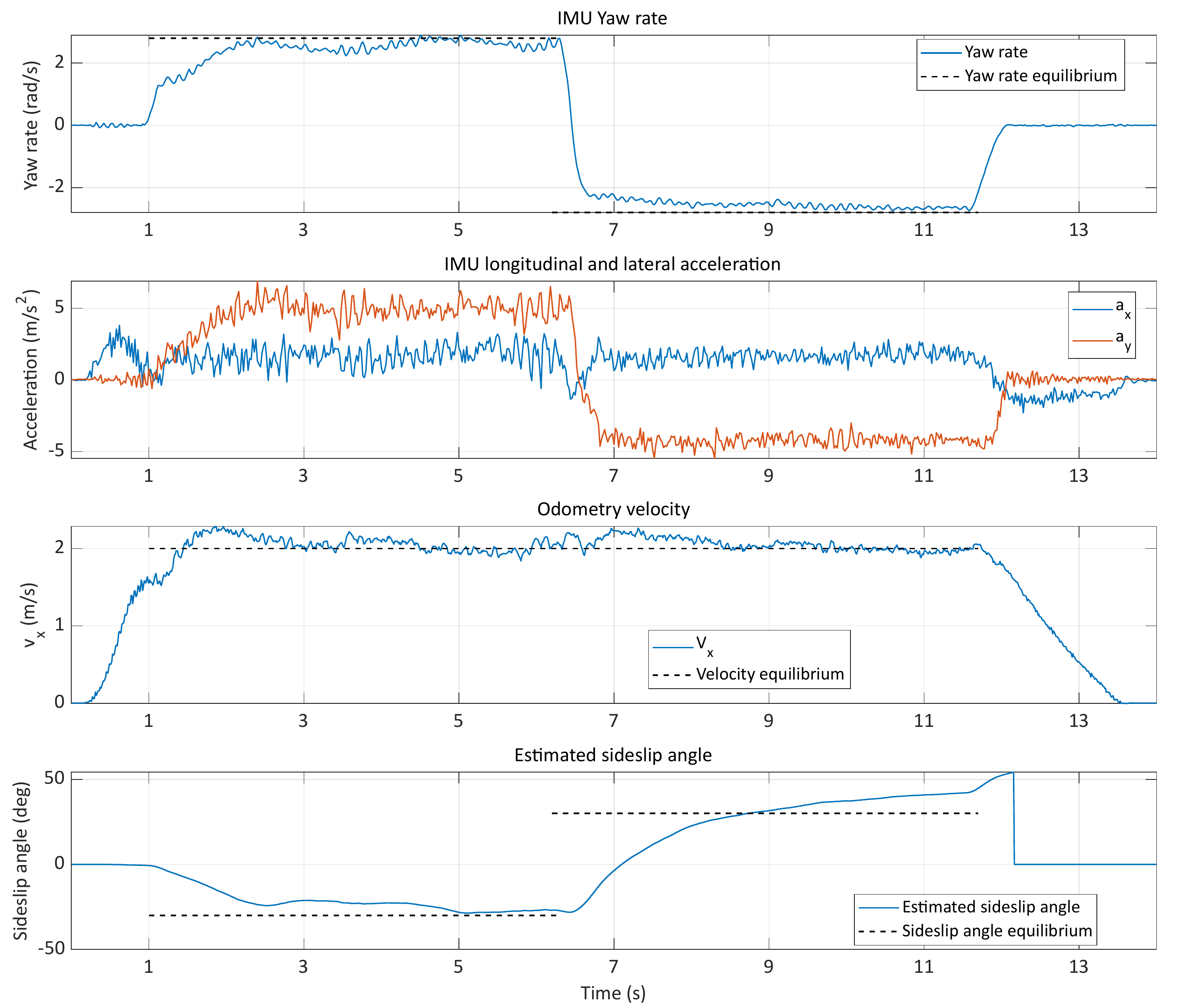}
\caption{Measured vehicle states during the figure-eight drift experiment, including the yaw rate, longitudinal and lateral accelerations, longitudinal velocity, and estimated sideslip angle.}
\label{fig:figure8_via}
\end{figure}

\begin{figure}[t]
\centering
\includegraphics[width=0.9\columnwidth]{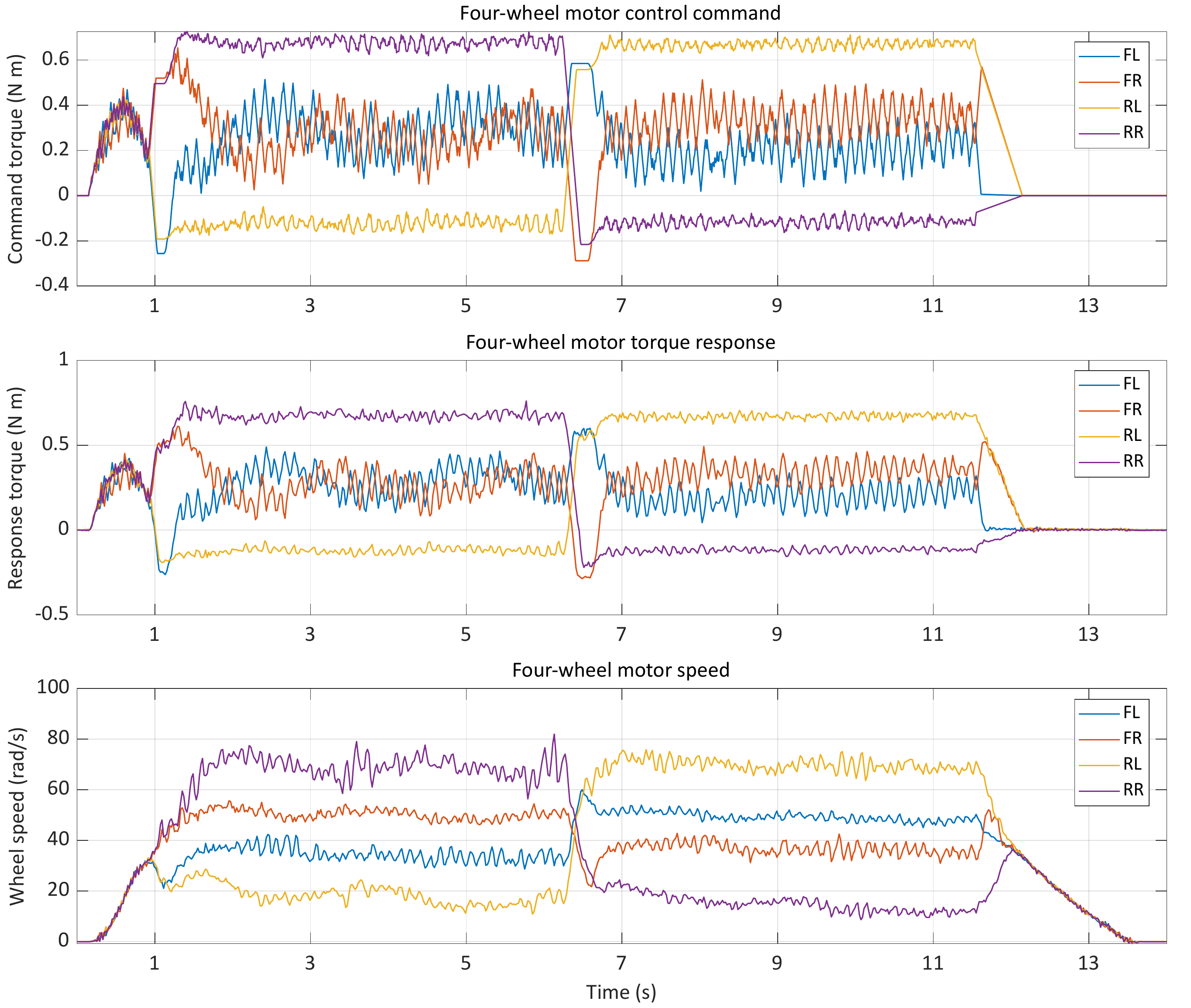}
\caption{Four-wheel torque commands, motor torque feedback, and wheel-speed responses during the figure-eight drift experiment.}
\label{fig:figure8_cmd}
\end{figure}

To further demonstrate the drifting capability of the steering-free 4WID chassis, a figure-eight drift experiment was conducted. In contrast to steady-state circular drifting, the figure-eight maneuver evaluates not only the tracking performance with respect to a prescribed drift equilibrium, but also the controller's ability to rapidly switch between two equilibria of opposite directions while maintaining continuous drift stabilization. During the figure-eight maneuver, the pulse and hold phases were each executed twice. The measured vehicle states are presented in Fig.~\ref{fig:figure8_via}, while the corresponding control commands and motor torque feedback are shown in Fig.~\ref{fig:figure8_cmd}.

For the figure-eight drift experiment, the desired longitudinal velocity was set to $2~\mathrm{m/s}$, and the magnitude of the equilibrium yaw rate was set to $2.8~\mathrm{rad/s}$. During the initial acceleration phase, the peak wheel torque was approximately $0.4~\mathrm{N,m}$, enabling the vehicle to approach the target velocity rapidly and smoothly. The vehicle then entered the first pulse phase, during which positive torques were applied to the front-right (FR) and rear-right (RR) wheels, while negative torques were applied to the front-left (FL) and rear-left (RL) wheels. The resulting differential torque generated the yaw moment required to drive the vehicle toward the first drift equilibrium.

During the subsequent hold phase, relatively large driving torques were maintained at the RL and RR wheels, whereas the FL and FR wheel torques were dynamically regulated to stabilize the vehicle around the desired drift equilibrium. The second pulse and hold phases followed the same control sequence, but with the wheel-torque directions reversed to establish and sustain drifting in the opposite direction, thereby completing the figure-eight maneuver.

The yaw-rate response during the first pulse phase was relatively slow, which may be associated with the selected torque magnitude during drift initiation. Nevertheless, the vehicle states subsequently converged toward the desired equilibrium, and a stable drifting posture was maintained throughout the hold phase. The experimental results therefore demonstrate that the proposed controller can both stabilize a prescribed drift equilibrium and perform rapid transitions between equilibria of opposite directions.

The vehicle sideslip angle $\beta$ was approximately estimated using the longitudinal and lateral accelerations measured by the IMU and the longitudinal velocity obtained from odometry. Based on the planar vehicle kinematics, the resultant velocity and sideslip-angle rate were computed as

\begin{equation}
\label{eq:beta_estimation}
\begin{aligned}
V &\approx \frac{u}{\cos\beta}, \
\dot{\beta}=\frac{a_y\cos\beta-a_x\sin\beta}{V}
-r,
\end{aligned}
\end{equation}

where $v_x$ denotes the longitudinal velocity, $a_x$ and $a_y$ are the longitudinal and lateral accelerations, respectively, $r$ denotes the yaw rate, and $V$ is the resultant velocity at the vehicle center of mass. The sideslip-angle estimate was subsequently obtained by numerically integrating $\dot{\beta}$. The resulting estimate is presented in Fig.~\ref{fig:figure8_via}.

The offline-computed drift equilibria correspond to sideslip-angle magnitudes of approximately $30^{\circ}$. As shown in Fig.~\ref{fig:figure8_via}, the estimated sideslip angle exhibits relatively small deviations from the equilibrium reference during the first drifting stage. During the second stage, however, the estimation accuracy deteriorates substantially. This degradation is primarily attributed to accumulated integration drift and the pronounced transient variations in the vehicle states.

\section{CONCLUSIONS}

This paper presented a differential-torque drift-control framework for a steering-free 4WID vehicle. A double-track dynamic model, a dedicated drift-equilibrium calculation method, and a pulse--hold controller were developed to generate and stabilize drift using only the four wheel torques. Simulations and experiments on a 1:10-scale platform demonstrated stable circular drifting with a sideslip angle of approximately $20^\circ$ and figure-eight drift tracking through transitions between equilibria of opposite directions. These results verify the feasibility of achieving near-limit drift motion without mechanical steering actuation.

The current approach relies on a nominal analytical vehicle model and has been validated under a limited set of operating conditions. Future work will employ neural networks to identify the real-vehicle dynamics from experimental data and thereby reduce modeling errors caused by unmodeled tire and chassis behavior. Reinforcement learning will also be investigated for drift control to improve generalization across vehicle speeds, tire--road friction conditions, and drift trajectories while preserving the physical constraints and stability requirements of the steering-free chassis.

\addtolength{\textheight}{-12cm}   







\bibliographystyle{IEEEtran}
\bibliography{main}

\end{document}